\documentclass[conference]{IEEEtran}
\IEEEoverridecommandlockouts
\usepackage{cite}
\usepackage{balance}
\usepackage{amsmath,amssymb,amsfonts}
\usepackage{algorithm}
\usepackage{algpseudocode}
\usepackage{graphicx}
\usepackage{multirow}
\usepackage{textcomp}
\usepackage[caption=false]{subfig}
\usepackage{adjustbox}
\usepackage[table,xcdraw]{xcolor}
\usepackage[utf8]{inputenc} 
\usepackage[T1]{fontenc} 
\def\BibTeX{{\rm B\kern-.05em{\sc i\kern-.025em b}\kern-.08em
    T\kern-.1667em\lower.7ex\hbox{E}\kern-.125emX}}

\newcommand{\dd}{\mathrm{\,d}}

\makeatletter
\def\BState{\State\hskip-\ALG@thistlm}
\makeatother

\usepackage{subfig}%
\usepackage{tikz}
\usepackage{pgfplots}
\tikzset{every picture/.append style={font=\scriptsize}}

\usepackage{siunitx}

\begin{document}

\title{Stochastic Optimization for Trajectory Planning with Heteroscedastic Gaussian Processes \\
}

\author{Luka Petrović, Juraj Peršić, Marija Seder, Ivan Marković$^\ast$
\thanks{
This research has been supported by the European Regional Development Fund under the grant KK.01.1.1.01.0009 (DATACROSS).
}
\thanks{
$^{\ast}$Authors are with the University of Zagreb
Faculty of Electrical Engineering and Computing, Laboratory for Autonomous Systems and Mobile Robotics, Croatia. {\{luka.petrovic, juraj.persic, marija.seder, ivan.markovic\}@fer.hr }
\newline 978-1-7281-3605-9/19/\$31.00 \textcopyright 2019 IEEE%
}
}

\maketitle

\begin{abstract}
Trajectory optimization methods for motion planning attempt to generate trajectories that minimize a suitable objective function.
Such methods efficiently find solutions even for high degree-of-freedom robots.
However, a globally optimal solution is often intractable in practice and state-of-the-art trajectory optimization methods are thus prone to local minima, especially in cluttered environments.
In this paper, we propose a novel motion planning algorithm that employs stochastic optimization based on the cross-entropy method in order to tackle the local minima problem.
We represent trajectories as samples from a continuous-time Gaussian process and introduce heteroscedasticity to generate powerful trajectory priors better suited for collision avoidance in motion planning problems.
Our experimental evaluation shows that the proposed approach yields a more thorough exploration of the solution space and a higher success rate in complex environments than a current Gaussian process based state-of-the-art trajectory optimization method, namely GPMP2, while having comparable execution time.
\end{abstract}

\begin{IEEEkeywords}
motion planning, trajectory optimization, gaussian processes, stochastic optimization
\vspace{-0.2cm}
\end{IEEEkeywords}

\section{Introduction}
\label{sec:intro}
Motion planning is an indispensable skill for robots that aspire to navigate through an environment without collisions.
Motion planning algorithms attempt to generate trajectories through the robot's configuration space that are both feasible and optimal based on some performance criterion dependent on the task, robot or environment.
Algorithms that can be executed in real time are highly encouraged, mostly because they allow fast replanning in response to environment changes.
The majority of methods in the domain of high-dimensional motion planning can be roughly divided into two categories: sampling-based approaches and trajectory optimization approaches.

The central tenet of sampling-based approaches \cite{kavraki1996probabilistic,lavalle2006planning, karaman2011sampling} is the idea of connecting points randomly sampled from the free configuration space.
Due to the underlying random sampling, these approaches exhibit probabilistic completness and fast exploration of the environment.
However, sampling based planners can be computationally inefficient for high-dimensional problems with challenging constraints and often require a post-processing step to smooth and shorten the computed trajectories.
Furthermore, considerable computational effort is spent on exploring the portions of the configuration space that might not be relevant to the task.

A significant amount of recent work has focused on trajectory optimization and related problems.
Trajectory optimization methods start with an initial trajectory and then minimize an objective function in order to optimize the trajectory.
Covariant Hamiltonian optimization for motion planning (CHOMP) \cite{chomp, chomp-ijrr} is the seminal work in modern trajectory optimization.
It utilizes a precomputed signed distance field for fast collision checking and uses covariant gradient descent to minimize obstacle and smoothness costs.
Stochastic trajectory optimization for motion planning (STOMP) algorithm \cite{stomp} samples a series of noisy trajectories to explore the space around an initial trajectory which are then combined to produce an updated trajectory with lower cost.
The key trait of STOMP is its ability to optimize non-differentiable constraints.
An important shortcoming of CHOMP and STOMP is the need for many trajectory states for reasoning about fine resolution obstacle representations and finding feasible solutions when there are many constraints.
TrajOpt \cite{trajopt}, \cite{schulman2014motion} algorithm formulates motion planning as sequential quadratic programming.
The key feature of TrajOpt is the ability to solve complex motion planning problems with few states since swept volumes are considered to ensure continuous-time safety.
However, if the smoothness is required in the output trajectory, either a densely parametrized trajectory or post-processing of the trajectory might still be needed thus increasing computation time.

The Gaussian process (GP) motion planning family of algorithms \cite{gpmp, gpmp2, gpmpgraph, gpmp-ijrr} employs continous-time trajectory representation in order to overcome the computational cost incurred by using large number of states.
The GPMP algorithm \cite{gpmp} parametrizes the trajectory with a few support states and then uses GP interpolation to query the trajectory at any time of interest.
The GPMP2 algorithm \cite{gpmp2} represents trajectories as samples from a continuous-time GP and then formulates the planning problem as probabilistic inference.
It exploits the sparsity of the underlying system by using preexisting optimization tools developed by the simultaneous localization and mapping (SLAM) community \cite{dellaert2012factor} to generate fast solutions.

Altough trajectory optimization methods generate fast solutions in high-dimensional spaces, they have limited exploration ability and in complex environments often converge to the infeasible local minima. In this paper, we propose a gradient-free stochastic optimization method for trajectory planning with continous time GP trajectory representations.
We consider a trajectory as a sample from a GP and introduce heteroscedasticity to generate powerful trajectory priors better suited for collision avoidance in motion planning problems.
The proposed optimization method relies on importance sampling and is a derivative of the cross-entropy optimization method \cite{rubinstein2013cross}.
While our method belongs to the trajectory optimization approaches, it relies on random trajectory samples which raises a connection to the sampling based planning.
The proposed method is an example of bridging the gap between sampling based and trajectory optimization approaches in order to generate fast solutions in high dimensional spaces while retaining the ability to throughly explore the environment.
We evaluated our method in simulations and compared it to GPMP2 -- a state-of-the-art gradient-based, in constrast to the proposed gradient-free, trajectory optimization method.
The results show that the proposed method yields a higher success rate in complex environments with comparable execution time.


\section{Heteroscedastic Gaussian Processes for Motion Planning}
\label{sec:gaussian}
\label{sec:gaussian}
\begin{figure*}[t]
     \centering
     \resizebox{.9\linewidth}{!}{
    \subfloat[Homoscedastic GP]{\includegraphics[width=0.34\textwidth]{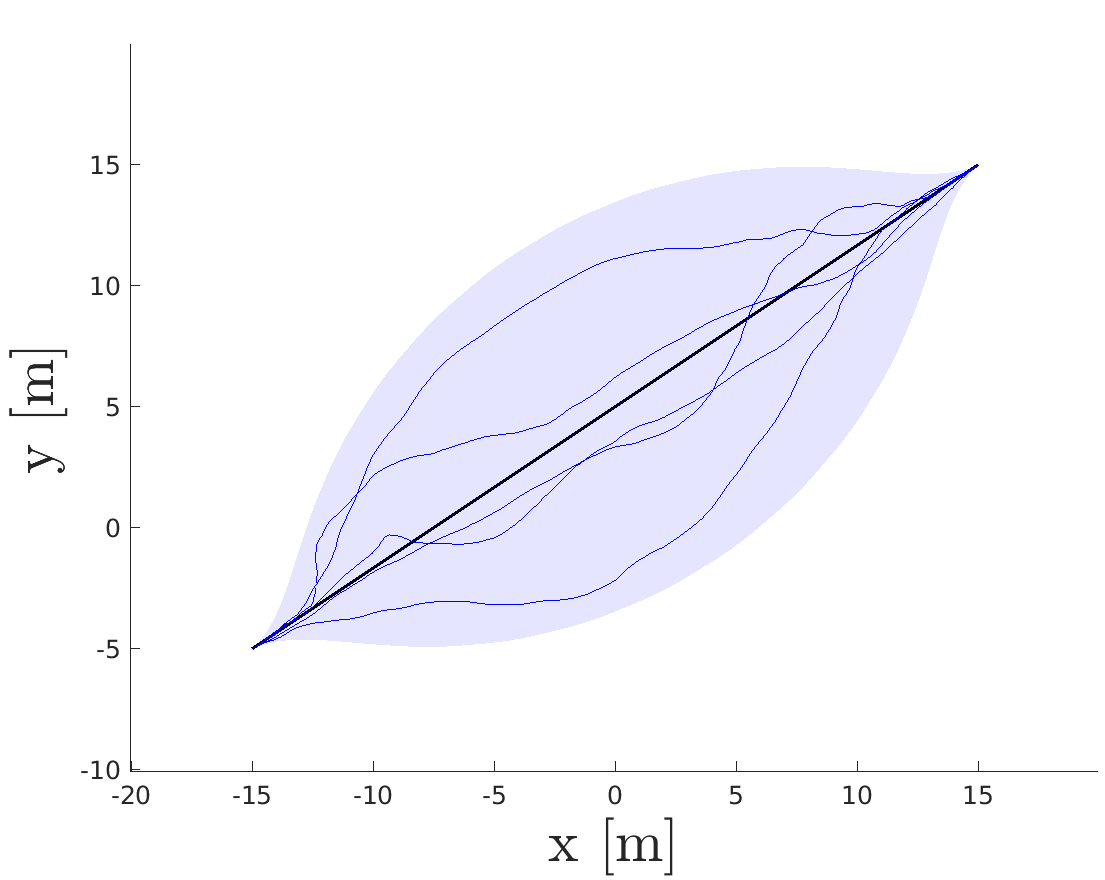}   \label{<figure2>}}
     \subfloat[Heteroscedastic GP]{\includegraphics[width=0.34\textwidth]{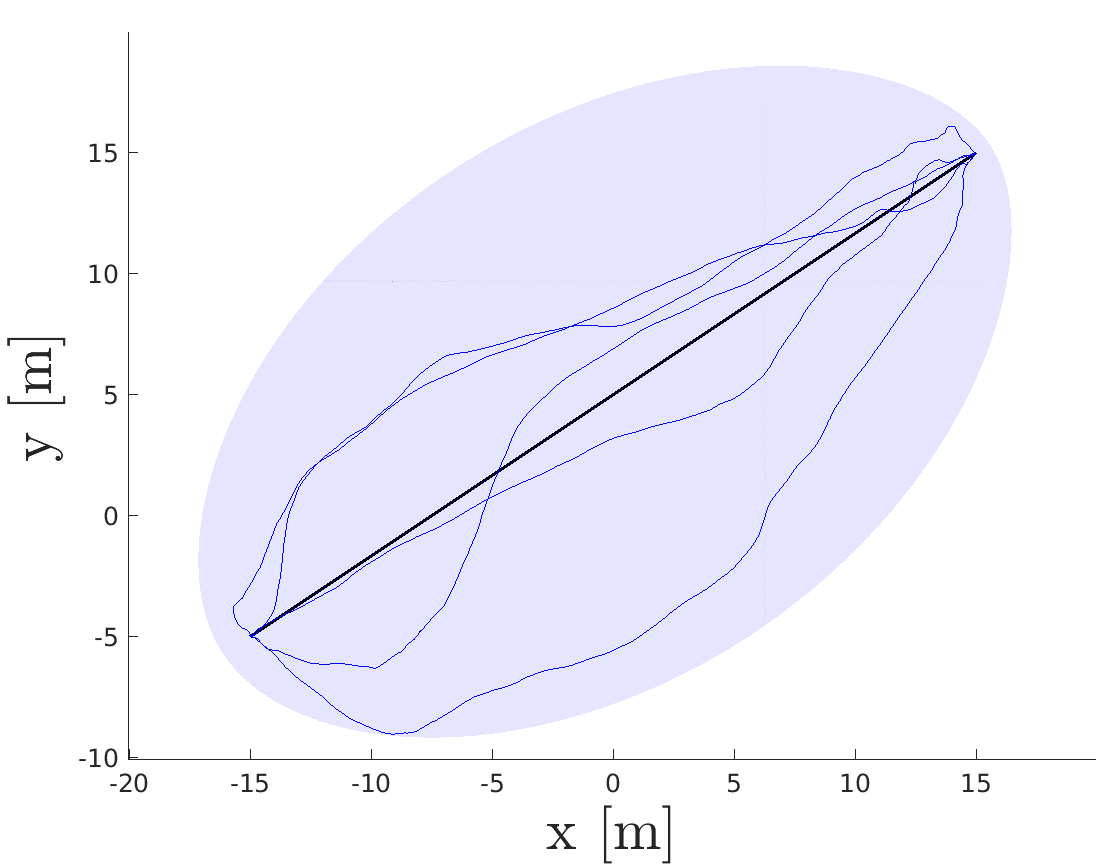}  \label{<figure1>}}
    \subfloat[Covariance parameter of the white noise process $\boldsymbol{w}(t)$ through time]{\adjustbox{raise=0.2pc}{\includegraphics[width=0.32\textwidth]{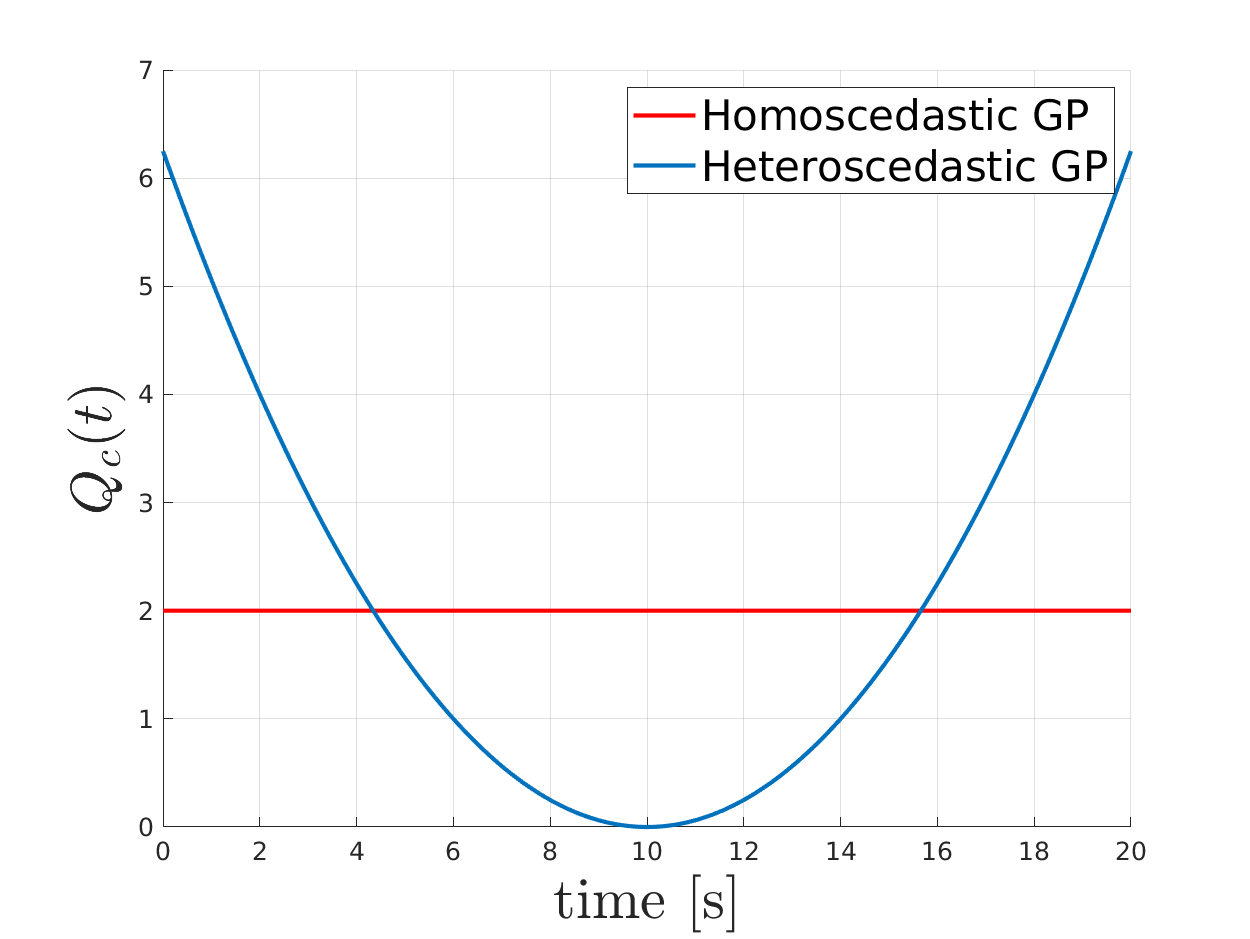} \label{<figure3>} }}
    }
     \caption{Comparison of the homoscedastic and the proposed heteroscedastic GP priors which depend on the covariance of the white noise process $\boldsymbol{w}(t)$}
     \label{fig:comparison}
     \vspace{-0.3cm}
\end{figure*}
\subsection{The Gaussian Process Trajectory Representations}
Consider a continuous-time trajectory as a sample from a vector-valued continuous-time Gaussian process (GP)
\begin{equation}
\boldsymbol{\theta}(t) \sim \mathcal{GP}(\boldsymbol{\mu}(t), \boldsymbol{\mathcal{K}}(t, t^{\prime}))
\label{eq:GPsample}
\end{equation}
that is parameterized with $N$ \textit{support states} at discrete time instants, $\boldsymbol{\theta}_i \in \mathrm{R}^D$, $i \in N$, where $D$ is the state dimensionality.
We employ a structured kernel belonging to a special class of GP priors generated by a linear time-varying stochastic differential equation (LTV-SDE)
%
%
\begin{equation}
\dot{\boldsymbol{\theta}}(t) = \boldsymbol{F}(t) \boldsymbol{\theta}(t) + \boldsymbol{v}(t) + \boldsymbol{L}(t) \boldsymbol{w}(t),
\label{ltvsde}
\end{equation}
where $\boldsymbol{F}$ and $\boldsymbol{L}$ are system matrices and $\boldsymbol{v}$ is a known exogenous input.
The white noise process $\boldsymbol{w}(t)$ is itself a GP with zero mean value
\begin{equation}
\boldsymbol{w}(t) \sim \mathcal{GP} (\boldsymbol{0}, \boldsymbol{Q}_c(t) \delta (t - t^{\prime})),
\end{equation}
where $\boldsymbol{Q}_c(t)$ is an isotropic time-varying power-spectral density matrix, $\boldsymbol{Q}_c(t) = Q_c(t) \boldsymbol{I}$.
A similar dynamical system has been utilized in estimation \cite{barfoot2014batch, anderson2015batch}, calibration \cite{Persic18preprint} and planning \cite{gpmp-ijrr, steap}. However, the crucial difference in our approach is that the covariance $\boldsymbol{Q}_c(t)$ is time-varying and consequently generates a heteroscedastic GP \cite{rana2017towards}. We discuss benefits of this approach in Section \ref{sec:heteroscedasticity}.

The mean and the covariance of the GP generated by the LTV-SDE given in \eqref{ltvsde} evaluate to
\begin{equation}
{\boldsymbol{\widetilde{\mu}}} (t) = \boldsymbol{\Phi}(t, t_0) {\boldsymbol{\mu}}_0 + \int_{t_0}^t \boldsymbol{\Phi} (t, s) \boldsymbol{v}(s) \dd s,
\label{prior_x}
\end{equation}
\begin{multline}
{\boldsymbol{\widetilde{\mathcal{K}}}}(t, t^{\prime}) =  \boldsymbol{\Phi}(t, t_0) {\boldsymbol{\mathcal{K}}}_{0} \boldsymbol{\Phi}(t^{\prime}, t_0)^T + \\ \int_{\text{t}_0}^{\text{min}(t,t^{\prime})} \boldsymbol{\Phi}(t, s)  \boldsymbol{L}(s) \boldsymbol{Q}_{c}(s) \boldsymbol{L}(s)^T \boldsymbol{\Phi}(t^{\prime}, s)^T \dd s,
\label{prior_P}
\end{multline}
where ${\boldsymbol{\mu}}_0$ and ${\boldsymbol{\mathcal{K}}}_0$ are the initial mean and covariance of the first state, and $\boldsymbol{\Phi} (t, s)$ is the state transition matrix \cite{barfoot2014batch}.
\subsection{GP Prior for Motion Planning}
Due to Markov property of the LTV-SDE in \eqref{ltvsde}, the inverse kernel matrix $\boldsymbol{\widetilde{\mathcal{K}}}^{-1}$ is exactly sparse block tridiagonal \cite{barfoot2014batch}:
\begin{equation}
\boldsymbol{\widetilde{\mathcal{K}}}^{-1} = \boldsymbol{\widetilde{F}}^{-T} \boldsymbol{\widetilde{Q}}^{-1} \boldsymbol{\widetilde{F}}^{-1},
\label{propagation}
\end{equation}
where
\begin{equation}
\boldsymbol{\widetilde{F}}^{-1} =
\begin{bmatrix}
\boldsymbol{1} & 0 & ... & 0 & 0 \\
-\boldsymbol{\Phi}(t_1, t_0) & \boldsymbol{1} & ... & 0 & 0 \\
0 & -\boldsymbol{\Phi}(t_2, t_1) & \ddots & \vdots & \vdots \\
\vdots & \vdots & \ddots & \boldsymbol{1} & 0 \\
0 & 0 & ... & -\boldsymbol{\Phi}(t_N, t_{N-1}) & \boldsymbol{1}
\end{bmatrix}
\end{equation}
and
\begin{equation}
\boldsymbol{\widetilde{Q}}^{-1} = \text{diag} ({\boldsymbol{\mathcal{K}}}_{0}^{-1}, \boldsymbol{Q}_{0,1}^{-1}, ... , \boldsymbol{Q}_{N-1, N}^{-1})
\end{equation}
with
\begin{equation}
\boldsymbol{Q}_{a, b} = \int_{t_a}^{t_b} \boldsymbol{\Phi}(t_b, s) \boldsymbol{L}(s) \boldsymbol{Q}_c \boldsymbol{L}(s)^T \boldsymbol{\Phi}(t_b, s)^T \dd s.
\label{eq:Qab}
\end{equation}

The GP defined by mean and covariance in \eqref{prior_x} and \eqref{prior_P} is well suited for estimation problems. However, in motion planning problems there exists a desired fixed goal state.
Given that, we need to condition this GP with a fictitious observation on the goal state with mean $\boldsymbol{\mu}_N$ and covariance $\boldsymbol{\mathcal{K}}_N$.
This can be accomplished while still preserving the sparsity of the kernel matrix \cite{gpmp-ijrr}
\begin{align}
	\boldsymbol{\mathcal{K}}^{-1} &= \small{\left[
    \begin{matrix}
      \widetilde{\boldsymbol{F}}^{-1} \\
      \begin{matrix}
        \boldsymbol 0 & \hdots & \boldsymbol 0 & \mathbf{I}
      \end{matrix}
    \end{matrix} \right]^\top
  \left[
    \begin{matrix}
      \widetilde{\boldsymbol{Q}}^{-1} & \nonumber \\
      & \boldsymbol{\mathcal{K}}_N^{-1}
    \end{matrix} \right]
  \left[
    \begin{matrix}
      \widetilde{\boldsymbol{F}}^{-1} \\
        \begin{matrix}
          \boldsymbol 0 & \hdots & \boldsymbol 0 & \mathbf{I}
        \end{matrix}
    \end{matrix} \right]}\\
	&= \boldsymbol{F}^T \boldsymbol{Q}^{-1} \boldsymbol{F}.
	\label{eq:sparseK}
\end{align}%
The mean vector $	\boldsymbol{\mu} = \begin{bmatrix} \boldsymbol{\mu}_0 \hdots \boldsymbol{\mu}_N \end{bmatrix}^T$ and the kernel matrix given in \eqref{eq:sparseK} fully determine a continuous-time trajectory defined by \eqref{eq:GPsample} that we employ for motion planning.
\subsection{Fast GP interpolation}
A major benefit of modelling continuous-time trajectory in motion planning with GPs is the possibility to query the planned state $\boldsymbol{\theta}(\tau)$ at any time of interest $\tau$, and not only at discrete time instants.
The kernel matrix defined in \eqref{eq:sparseK} allows for computationally efficient, structure-exploiting GP interpolation with $\mathcal{O}(1)$ complexity.
State $\boldsymbol{\theta}(\tau)$ at $\tau \in [t_i, t_{i+1}]$ is a function only of its neighboring states~\cite{gpmp2}
\begin{equation}
{\boldsymbol{\theta}}(\tau) = {\boldsymbol{\mu}}(\tau) + \boldsymbol{\Lambda}(\tau)({\boldsymbol{\theta}}_i - {\boldsymbol{\mu}}_i) + \boldsymbol{\Psi}(\tau)({\boldsymbol{\theta}}_{i+1} - {\boldsymbol{\mu}}_{i+1}),
\label{eq:intp1}
\end{equation}
\begin{equation}
\boldsymbol{\Lambda}(\tau) = \boldsymbol{\Phi} (\tau, t_i) - \boldsymbol{\Psi}(\tau)\boldsymbol{\Phi}(t_{i+1}, t_i),
\label{eq:intp2}
\end{equation}
\begin{equation}
\boldsymbol{\Psi}(\tau) = \boldsymbol{Q}_{i, \tau} \boldsymbol{\Phi}(t_{i+1}, \tau)^T \boldsymbol{Q}_{i, i+1}^{-1},
\label{eq:intp3}
\end{equation}
where $\boldsymbol{Q}_{a,b}$ is given in \eqref{eq:Qab}.
Efficient GP interpolation can be exploited for reasoning about small obstacles while keeping a relatively small number of \textit{support states} which reduces the incurred computational burden.
It can also be utilized for providing a dense output trajectory that a robot can execute without any post-processing.
\subsection{Constant-velocity motion model}
Robot dynamics are represented with the double integrator linear system with white noise injected in acceleration. The trajectory is thus generated by the LTV-SDE \eqref{ltvsde}, where the Markovian state $\boldsymbol{\theta}(t)$ consists of position and velocity in configuration space with the following system matrices
\begin{equation}
\boldsymbol{F}(t) = \begin{bmatrix}
\boldsymbol{0} & \textbf{I} \\
\boldsymbol{0} & \boldsymbol{0}
\end{bmatrix}, \, \boldsymbol{L}(t) = \begin{bmatrix}
\boldsymbol{0} \\ \textbf{I}
\end{bmatrix}.
\label{eq:constvel}
\end{equation}
This formulation generates a constant velocity GP prior which is centered around a zero acceleration trajectory. Applying this motion model minimizes actuator acceleration in the configuration space, thus minimizing the energy consumption and providing the physical meaning of smoothness \cite{gpmp-ijrr}.
\subsection{Benefits of heteroscedasticity}
\label{sec:heteroscedasticity}
State-of-the-art methods based on GPs \cite{gpmp2, gpmpgraph, gpmp-ijrr} minimize the sum of two costs: an obstacle cost and a smoothness cost which measures the deviation of the trajectory from the GP prior mean.
Those methods use covariance as a parameter in optimization, with smaller values of $\boldsymbol{\mathcal{K}}$ penalizing the deviation of the trajectory from the prior mean more.
For therein employed homoscedastic GPs, small constant values of $\boldsymbol{Q}_c$ allow high variance of states near the midpoint of trajectory, while states close to the trajectory start or goal have relatively small variance.
If the prior mean of those states passes through an obstacle, the Levenberg-Marquardt optimization technique used in \cite{gpmp2, gpmpgraph, gpmp-ijrr} will have difficulty escaping the collision since the incurred smoothness cost can become too large commensurate with obstacle cost.
Figure \ref{<figure2>} shows that sample trajectories drawn from the homoscedastic GP  do not have large deviation from the mean near the first and  the last state.
With larger constant values of $\boldsymbol{Q}_c$ this problem diminishes, however, states near the trajectory midpoint then have high variance and trajectories lose the desirable smoothness property.

By introducing heteroscedasticity of the underlying white noise process, we can design GPs that are better suited for motion planning and help alleviate the problem of local minima.
Ideally, the GP should be able to generate trajectories that maneuver around the obstacles near start and goal states, while retaining the smoothness property in the middle.
With careful selection of the proposed time-varying white noise power-spectral density matrix $\boldsymbol{Q}_c(t)$, we are able to model GPs that achieve the stated goal.
In our experience, modeling $Q_c(t)$ as a parabola with high values at the beginning and end, and the lowest value at the temporal midpoint of the trajectory, leads to GPs that have these desirable features.
Note that the GP covariance $\boldsymbol{\mathcal{K}}(t, t^{\prime})$ is obtained by propagating $Q_c(t)$ with the underlying motion model, as defined in \eqref{propagation}, and thus low (or even zero) value of $Q_c(t)$ at a particular time instant does not imply small GP covariance.
An example of such $Q_c(t)$ is depicted in Figure~\ref{<figure3>} and it leads to a heteroscedastic GP shown in Figure~\ref{<figure1>}.
Notably, sample trajectories drawn from the example heteroscedastic GP have large deviation from the mean near the first and the last state.
The example GP would be able to thoroughly explore the environment and generate trajectories that bypass obstacles near the first and the last state.
Note that any non-negative function can be used for $Q_c(t)$, depending on the specific context of some motion planning problem.
For example, one could model $Q_c(t)$ as a monotonically decreasing function, resembling the aim of exploring more at the beginning and less towards the end.

\section{Proposed Stochastic Trajectory Optimization}
\label{sec:method}
Formally, the goal of trajectory optimization is to find a smooth, collision-free trajectory through the configuration space between two end points. Prior work in this area models the cost of a trajectory using two terms: a prior term, which usually encodes smoothness that minimizes higher-order derivatives of the robot states, and an arbitrary state-dependent term, which usually measures the cost of being near obstacles. However, our optimization criteria consists solely of an arbitrary state-dependent cost term. We reason that our trajectory carries an inherent property of smoothness since we model it as a sample from the GP defined in  \eqref{eq:GPsample}. Therefore, our method starts with the following optimization problem:
\begin{equation}
\begin{aligned}
&  \underset{\boldsymbol{\theta}(t)}{\text{minimize}}
&& f [\boldsymbol{\theta}(t)] \\
& \text{subject to}
&&  \boldsymbol{\theta}(t) \sim \mathcal{GP}(\boldsymbol{\mu}(t), \boldsymbol{\mathcal{K}}(t, t^{\prime})). \\
\end{aligned}
\label{eq:optimcost}
\end{equation}
The state-dependent cost term $f [\boldsymbol{\theta}(t)]$ can include any cost function corresponding to the desired trajectory properties, e.g. collision avoidance, task-space constraints, torques \cite{stomp} and manipulability \cite{maric2018manipulability}.
In this work, we consider only collision avoidance and use a precomputed signed distance field for collision checking similarly to \cite{gpmp2, gpmp-ijrr}.


Most of the state-of-the-art approaches use gradient-based methods which find locally optimal trajectories.
In this work, we instead optimize using a derivative-free stochastic optimization method. This allows for better exploration while being less prone to local minima.
It also enables optimization of arbitrary costs which are non-differentiable or non-smooth.
To solve \eqref{eq:optimcost}, we employ a stochastic optimization approach stemming from the cross-entropy method \cite{rubinstein2013cross} and with similarities to the estimation-of-distribution algorithm \cite{larranaga2001estimation}.

Our method starts with drawing $K$ sample trajectories from the GP defined in  \eqref{eq:GPsample}, where the mean $\boldsymbol{\mu}$ is initialized as a constant-velocity straight line in configuration space and covariance matrix $\boldsymbol{\mathcal{K}}$ arises from the kernel matrix defined in  \eqref{eq:sparseK}.
A sample trajectory is generated using
\begin{equation}
  \boldsymbol{\theta} = \boldsymbol{\mu} + \boldsymbol{A}\boldsymbol{Z},
\end{equation}
where $\boldsymbol{A}$ is a lower triangular matrix obtained by Cholesky decomposition of the covariance matrix, $\boldsymbol{\mathcal{K}} = \boldsymbol{A} \boldsymbol{A}^T$, and $\boldsymbol{Z}$ is a vector of $N$ standard normal variables $\boldsymbol{Z} \sim \mathcal{N}(\boldsymbol{0}, \textbf{I})$.

Subsequently, we evaluate the cost $f [\boldsymbol{\theta}(t)]$ for each trajectory using the aforementioned hinge loss with a precomputed signed distance field.
From the evaluated $K$ trajectories we pick $M$ best ones according to the optimization criteria, i.e. ones with the lowest cost $f [\boldsymbol{\theta}(t)]$.
We then take the cost-weighted average (weighted arithmetic mean) of those $M$ best trajectories in order to form a GP mean $\boldsymbol{\mu}$ for next iteration:
\begin{equation}
  \boldsymbol{\mu}(t) = \frac {\sum_{m=1}^{M} w_m \boldsymbol{\theta}_m (t)}{ \sum_{m=1}^{M} w_m}
\end{equation}
where
\begin{equation}
  w_m = 1 / f[\boldsymbol{\theta}_m(t)].
\end{equation}
This process is repeated until a collision-free trajectory is found. The described method is summarized in Algorithm~\ref{algoritam}.

Contrary to a generic cross-entropy optimization algorithm, in our approach the covariance matrix $\boldsymbol{\mathcal{K}}$ remains unchanged through iterations.
This is done because computing the new covariance matrix of the same form would significantly increase the computational burden.
Furthermore, the unchanged covariance matrix allows for exhaustive exploration around the mean in each iteration.
Changing only mean $\boldsymbol{\mu}$ while keeping the covariance matrix $\boldsymbol{\mathcal{K}}$ unchanged is permitted in the GP framework described in Section \ref{sec:gaussian}, as change in $\boldsymbol{\mu}$ can be attributed to some implicitly imposed exogenous input $\boldsymbol{v}(t)$ which does not impact covariance.


\begin{algorithm}
\caption{Stochastic Trajectory Optimization with GPs}\label{algoritam}
\begin{algorithmic}[1]
\Require { Start and goal states $\theta_0$, $\theta_N$, a state-dependent cost function $f[\boldsymbol{\theta}_k(t)]$}
\Ensure{Initial mean $\boldsymbol{\mu}$ and covariance $\boldsymbol{\mathcal{K}}$}
\For {$1 \dots N_{iter}$}
\For {$1 \dots K$}
\State Sample trajectory $\boldsymbol{\theta}_k(t) \sim \mathcal{GP}(\boldsymbol{\mu}(t), \boldsymbol{\mathcal{K}}(t, t^{\prime}))$
\State Evaluate trajectory cost $f [\boldsymbol{\theta}_k(t)]$
\If {$f [\boldsymbol{\theta}_k(t)] = 0$}
\State Return collision free trajectory $\theta_k(t)$
\EndIf
\EndFor
\State From K sampled trajectories take M with lowest cost
\State Compute new mean $\boldsymbol{\mu}(t) = \frac {\sum_{m=1}^{M} [w_m \boldsymbol{\theta}_m (t)]}{ \sum_{m=1}^{M} w_m} $
\EndFor
\end{algorithmic}
\end{algorithm}
\subsection{Computational Efficiency Remarks}
In order to throughly explore the environment, our approach requires cost evaluation for relatively many drawn trajectory samples, which naturally leads to the slower computation than the state-of-the-art gradient based methods.
However, due to the fact that cost evaluation for each trajectory is independent, the inner for loop in the proposed Algorithm~\ref{algoritam} can be parallelized with computational efficiency scaling linearly with the number of processing cores.
In our implementation, we exploit this property and parallelize the inner loop on 4 processing cores.
A GPU implementation presents an interesting possibility that would allow sampling and cost evaluation for a huge number of trajectories, leading to fast environment exploration and discovering optimal trajectories allowing real-time replanning in dynamic environments \cite{park2013real}.

We use GP interpolation for dense collision checking, similarly to \cite{gpmp2}.
Since trajectory \textit{support states} are temporally equidistant and each sampled trajectory is drawn from the same GP, matrices $\Lambda$ and $\Psi$ in \eqref{eq:intp2} and \eqref{eq:intp3} can be precomputed, instead of computing them each time interpolation is needed.
This provides another significant increase of the proposed method computational efficiency.


\section{Test Results}
\label{sec:results}
We tested the proposed method on two simulation benchmarks and compared it with the state-of-the-art trajectory optimization technique GPMP2 \cite{gpmp2}.
In Section~\ref{sec:maze}, we quantitatively demonstrate the improvement of the proposed method over GPMP2 with random restarts in solving a 2D maze, which is a good benchmark for an optimization-based planner effectiveness at finding a collision-free solution in a haystack of local minima.
This experiment aims to show benefits of the proposed stochastic method, which allows for better exploration in comparison to gradient-based methods.
In Section~\ref{sec:manipulation}, we demonstrate the improvement of the proposed method over prior techniques in finding a collision-free trajectory for a $7$\,DOF manipulator in cluttered environment.
This experiment aims to show benefits of the proposed heteroscedastic prior since the environment was set up so that obstacles are placed near the start and goal state.

In both benchmarks, our method was always initialized with a constant-velocity straight line trajectory in the  configuration space.
For GPMP2, we used a straight line initialization as a baseline, and in our experiments we designate to this model as  \textit{line}.
Since GPMP2 always converges very quickly, but often fails in cluttered environments due to infeasible local minima, we also employed random restarts, which is a commonly used method to tackle the local minima problems in gradient-based trajectory optimization methods \cite{chomp}.
In this technique, the optimizer is first initialized with a straight-line and, on failure, re-initialized with a random trajectory.
Our implementation samples the random restart trajectory from a homoscedastic GP, similarly to \cite{gpmpgraph}.
We designate to this model as \textit{rr}.

We used the GPMP2 C++ library \cite{gpmp2, dong2017sparse} and its respective MATLAB toolbox based on the GTSAM C++ library \cite{dellaert2012factor}.
Experiments were performed on a system with a 2.8-GHz Intel Core i7-7700HQ processor and 16 GB of RAM.

\subsection{The Maze Benchmark}
\label{sec:maze}
The maze benchmark, appropriate for quantitative evaluation, consisted of 1000 synthetic environments created by the Wilson's algorithm \cite{wilson1996generating}, which generates uniformly sampled mazes with a single solution (i.e. perfect mazes).
Mazes were generated on grids with sizes of $3\times3$, $4\times4$ and $5\times5$ and afterwards inflated to realistic dimensions.
While the 2D maze problem is generally suitable for grid-based or sampling-based motion planning approaches which achieve $100$$\%$ success rate, it can be used to measure an optimization-based planner's effectiveness at finding the unique collision-free solution in a cluttered environment.

For each maze environment, we plan motion for a 2D holonomic circular robot with the radius of $0.5$\,m.
For our method, we chose the number of sampled trajectories $K \in [200, 400]$, while the number of best trajectories chosen for the weighted mean in each iteration was set to $M = 3$.
Altough these parameters may seem disproportionate, choosing a large $K$ ensures exhaustive exploration, while small $M$ induces drastic changes in the GP mean $\boldsymbol{\mu}$ between iterations, which helps in finding the solution faster in complex environments.
We set the total trajectory time (i.e. the timespan in which robot moves from start to goal state) to $t_{\mathrm{total}} = 20$\,s, while time-varying covariance matrix of the white noise governing the heteroscedastic GP was calculated as $\boldsymbol{Q}_c(t) = (t - \frac{t_{\mathrm{total}}}{2})^2$, which generates a parabola with its vertex at the midpoint of the trajectory.
For GPMP2 we used the parameters set from the Matlab toolbox 2D example.
For both methods, the trajectory was parametrized with $N = 10$ \textit{support states} and $5$ interpolation steps inbetween for which the trajectory cost is evaluated.
We set the maximum runtime for our algorithm and random restarts as $t_{\mathrm{max}} = 1$\,s, with one exception where we set $t_{\mathrm{max}} = 2$\,s in order to investigate the ability of our algorithm to find solutions given more time.
We measure the number of mazes solved (success rate) and the execution time.
The reults of the experiment are shown in Table~\ref{tab:maze}.

\begin{figure}[t]
     \centering
    \includegraphics[width=0.35\textwidth]{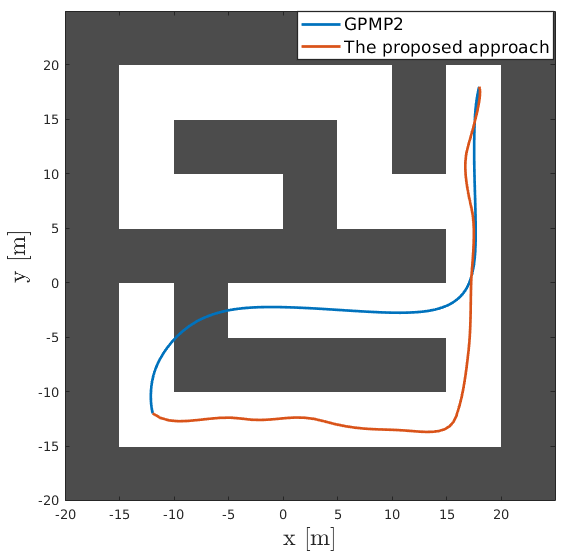}
     \caption{Example of a $4\times4$ maze where the proposed approach finds a collision-free solution, while GPMP2 converges to an infeasible local minimum. Slight undulation of the trajectory obtained by the proposed method is due to the criterium of finding a collision-free trajectory, unlike the GPMP2 criterium which explicitly encourages smoothness.}
     \label{fig:maze}
     \vspace{-0.2cm}
\end{figure}

While the GPMP2 without random restarts has an order of magnitude faster execution time, it has the worst success rate for every maze complexity level.
For the least complex mazes, created from a $3\times3$ grid, the random restarts outperformed our algorithm, managing similarly high success rate with significantly faster computation.
However, for the mazes created from a $4\times4$ grid our algorithm outperformed random restarts, having notably higher success rate with similar reported times.
Note that the reported execution time for our algorithm is the actual time it took to compute, and not the sum over all cores.
An example of a $4\times4$ maze is shown in Figure~\ref{fig:maze}.
The most complex mazes created from a $5\times5$ grid demonstrated the inability of gradient-based methods to find solutions in complex environments plagued with multitude of local minima.
\begin{table}[]
\centering
\caption{Success rate (percentage) / average execution time (miliseconds) on maze and robot arm planning benchmarks.}
\begin{adjustbox}{max width=0.47\textwidth}
\label{tab:maze}
\begin{tabular}{cccccc}
\hline
                      & \multicolumn{3}{c}{The proposed approach} & \multicolumn{2}{c}{GPMP2}                  \\
                      \cline{2-6}
\multirow{2}{*}{Maze} & $K = 400$        & $K = 400$         & $K = 200$        & \multirow{2}{*}{rr} & \multirow{2}{*}{line} \\
                      & $t_{max} = 2$\,s & $t_{max} = 1$\,s  & $t_{max} = 1$\,s &                     &                       \\ \hline
3x3                   & 95.2 / 197       & 92.9 / 171        & 89.0 / 160       & 89.2 / 96           & 55.3 / 28             \\
4x4                   & 79.1 / 489       & 66.9 / 324        & 61.8 / 297       & 44.5 / 301          & 13.8 / 39             \\
5x5                   & 43.8 / 858       & 26.7 / 493        & 26.3 / 429       & 5 / 252             & 2.1 / 29               \\ \hline
\hline
\multirow{2}{*}{Arm} & & Heteroscedastic       & Homoscedastic       & rr           & line        \\ \cline{3-6}
& & 100 / 446             & 75 / 548            & 80 / 368     & 10 / 45     \\ \hline
\end{tabular}
\end{adjustbox}
\vspace{-0.3cm}
\end{table}
\subsection{The Robot Arm Planning Benchmark}
\label{sec:manipulation}
The robot arm planning benchmark consisted of a simulated WAM robotic arm in an environment featuring a table and a drawer.
We conducted $20$ unique experiments, all with different start and goal states with starting points being under the table and end states being inside the drawer.
This set of problems is not particularly difficult since most of the states are initially collision free, however it was set up to accentuate the proneness of the homoscedastic GP planning methods to get stuck in local minima near start or goal states.

For our method, we chose the number of sampled trajectories $K = 400$, while the number of best trajectories chosen for the weighted mean in each iteration was set as $M = 3$.
In this benchmark we tested our optimization method with both heteroscedastic and homoscedastic GP priors in order to demonstrate the benefits of heteroscedasticity.
We set the total trajectory time $t_{\mathrm{total}} = 20$\,s.
For homoscedastic case we chose $\boldsymbol{Q}_c = 2$, while for a heteroscedastic GP we calculated $\boldsymbol{Q}_c(t) = (t - \frac{t_{\mathrm{total}}}{2})^2$, similarly to the maze benchmark.
For GPMP2 we used the default parameters set up in the Matlab toolbox WAM planner example.
For both methods the trajectory was parametrized with $N = 10$ equidistant \textit{support states} and $10$ interpolation steps inbetween for which the trajectory cost was evaluated.
We again set the fixed time budget for our algorithm and random restarts as $t_{\mathrm{max}} = 1$\,s.
We again measured the success rate and the execution time.
The results of the experiment are shown in Table~\ref{tab:maze}, where we can see that while the baseline GPMP2 fails in most cases, the stochasticity introduced by random restarts helps in achieving higher success rates.
The proposed method achieved a perfect score within the fixed time budget, thus demonstrating the advantage of the proposed heteroscedastic prior.
\begin{figure}[t]
     \centering
    \includegraphics[width=0.44\textwidth]{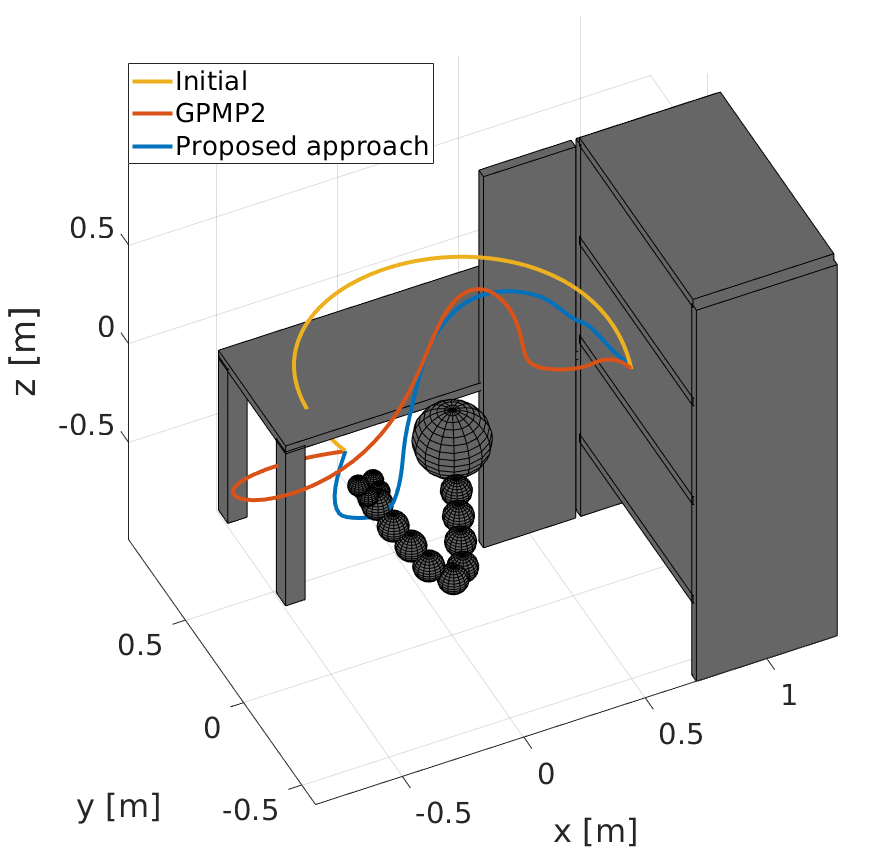}
     \caption{A simulated WAM robotic arm in an environment featuring a table and a drawer. Plotted lines depict the end effector trajectories. This is an example where the proposed approach finds a collision free solution, while GPMP2 converges to the infeasible local minimum. Initial straight-line trajectory in configuration space is also shown.}
     \label{fig:manipulation}
     \vspace{-0.25cm}
\end{figure}

\section{Conclusion}
\label{sec:conclusion}
In this paper we have presented a stochastic trajectory optimization method for motion planning. We considered each trajectory as a sample from a continuous time GP generated by a linear time-varying stochastic differential equation. By introducing the heteroscedasticity of the underlying GP, we were able to generate trajectory priors better suited for collision avoidance in motion planning problems. We proposed a cross-entropy method based derivative-free optimization in order to contend with the local minima problem present in trajectory optimization methods. Through simulated experiments we demonstrated that the proposed approach ameliorated the local minima problem present in trajetory optimization approaches while having comparable execution time.

In future work, it would be interesting to exploit the parallelization capability of our algorithm with a GPU implementation.
Furthermore, the strong exploration capability could be used for finding homotopy classes in the environment.
\balance

\bibliographystyle{IEEEtran}
\bibliography{main}

\end{document}